\documentclass[letterpaper, 10 pt, conference]{ieeeconf}  % Comment this line out
\usepackage{graphicx}
\IEEEoverridecommandlockouts                              % This command is only
\title{\LARGE \bf Personalized Taste and Cuisine Preference Modeling via Images}

\author{Nitish Nag$^{1,3}$, Bindu Rajanna$^{2}$, Ramesh Jain$^{1}$% <-this % stops a space
\thanks{*This work was not supported by any organization}% <-this % stops a space
\thanks{$^{1}$ Donald Bren School of Information and Computer Science,
        University of California, Irvine, United States of America
        {\tt\small  jain at ics.uci.edu}}%
\thanks{$^{2}$ Department of Computer Science, PES University,
        Bangalore, India
        {\tt\small bindu.rajanna539@gmail.com}}%
 \thanks{$^{3}$ Medical Scientist Training Program, School of Medicine, University of California, Irvine, United States of America,
        {\tt\small nagn at uci.edu}}%
}

\begin{document}

\maketitle
\thispagestyle{empty}
\pagestyle{empty}

%%%%%%%%%%%%%%%%%%%%%%%%%%%%%%%%%%%%%%%%%%%%%%%%%%%%%%%%%%%%%%%%%%%%%%%%%%%%%%%%
\begin{abstract}

With the exponential growth in the usage of social media to share live updates about life, taking pictures has become an unavoidable phenomenon. Individuals unknowingly create a unique knowledge base with these images. The food images, in particular, are of interest as they contain a plethora of information. From the image metadata and using computer vision tools, we can extract distinct insights for each user to build a personal profile. Using the underlying connection between cuisines and their inherent tastes, we attempt to build such a profile for an individual based solely on the images of his food. Our study provides insights about an individual's inclination towards a certain cuisine. Interpreting these insights can lead to the development of a more precise recommendation system. Such a system would avoid the generic approach in favor of a personalized system.

\end{abstract}

%%%%%%%%%%%%%%%%%%%%%%%%%%%%%%%%%%%%%%%%%%%%%%%%%%%%%%%%%%%%%%%%%%%%%%%%%%%%%%%%
\section{INTRODUCTION}

A picture is worth a thousand words. Complex ideas can easily be depicted via an image. An image is a mine of data in the 21st century. With each person taking an average of 20 photographs every day, the number of photographs taken around the world each year is astounding. According to a Statista report on Photographs, an estimated 1.2 trillion photographs were taken in 2017 and 85\% of those images were of food. Youngsters can't resist taking drool-worthy pictures of their food before tucking in. Food and photography have been amalgamated into a creative art form where even the humble home cooked meal must be captured in the perfect lighting and in the right angle before digging in. According to a YouGov poll, half of Americans take pictures of their food.

The sophistication of smart-phone cameras allows users to capture high quality images on their hand held device. Paired with the increasing popularity of social media platforms such as Facebook and Instagram, it makes sharing of photographs much easier than with the use of a standalone camera. Thus, each individual knowingly or unknowingly creates a food log. 

A number of applications such as MyFitnessPal, help keep track of a user's food consumption. These applications are heavily dependent on user input after every meal or snack. They often include several data fields that have to be manually filled by the user. This tedious process discourages most users, resulting in a sparse record of their food intake over time. Eventually, this data is not usable. On the other hand, taking a picture of your meal or snack is an effortless exercise. 

Food images may not give us an insight into the quantity or quality of food consumed by the individual but it can tell us what he/she prefers to eat or likes to eat.  We try to tackle the following research question with our work:
Can we predict the cuisine of a food item based on just it's picture, with no additional text input from the user? 

\section{RELATED WORK}

The work in this field has not delved into extracting any information from food pictures. The starting point for most of the research is a knowledge base of recipes (which detail the ingredients) mapped to a particular cuisine. 

Han Su et. al.\cite{Su2014AutomaticIngredients} have worked on investigating if the recipe
cuisines can be predicted from the ingredients of recipes. They treat ingredients as features and provide insights on which cuisines are most similar to each other. Finding common ingredients for each cuisine is also an important aspect. Ueda et al. \cite{UedaRecipeRecipe} \cite{UedaUsersRecommendation} proposed a personalized recipe recommendation method based on users' food preferences. This is derived from his/her recipe browsing activities and cooking history. 

Yang et al \cite{ShulinYang2010FoodFeatures} believed the key to recognizing food is exploiting the spatial relationships between different ingredients (such as meat and bread in a sandwich). They propose a new representation for food items that calculates pairwise statistics between local features computed over a soft pixel-level segmentation of the image into eight ingredient types. Then they accumulate these statistics in a multi-dimensional histogram, which is then used as a feature vector for a discriminative classifier.

Existence of huge cultural diffusion among cuisines is shown by the work carried out by S Jayaraman et al in \cite{Jayaraman2017AnalysisLearning}. They explore the performance of each classifier for a given type of dataset under unsupervised learning methods(Linear support Vector Classifier (SVC), Logistic Regression, Random Forest Classifier and Naive Bayes). 

H Holste et al's work \cite{Holste2015WhatRecipes} predicts the cuisine of a recipe given the list of ingredients. They eliminate distribution of ingredients per recipe as a weak feature. They focus on showing the difference in performance of models with and without tf-idf scoring. Their custom tf-idf scoring model performs well on the Yummly Dataset but is considerably naive.

R M Kumar et al \cite{RahulVenkateshKumar2016CuisineAlgorithms} use Tree Boosting algorithms(Extreme Boost and Random Forest) to predict cuisine based on ingredients. It is seen from their work that Extreme Boost performs better than Random Forest.

Teng et al \cite{Teng2011RecipeNetworks} have studied substitutable ingredients using recipe reviews by creating substitute ingredient graphs and forming clusters of such ingredients.

\section{DATASET}
The Yummly\cite{GauravSood2017Clarifai:API} dataset is used to understand how ingredients can be used to determine the cuisine. The dataset consists of 39,774 recipes. Each recipe is associated with a particular cuisine and a particular set of ingredients. Initial analysis of the data-set revealed a total of 20 different cuisines and 6714 different ingredients. Italian cuisine, with 7383 recipes, overshadows the dataset.

The numbers of recipes for the 19 cuisines is quite imbalanced.\cite{Ahn2011FlavorPairing} The following graph shows the count of recipes per cuisine.

\begin{figure}[thpb]
      \centering
        \includegraphics[width=0.5\textwidth]{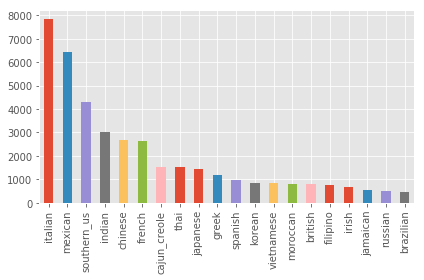}
      \caption{ Count of Recipes per Cuisine}
\end{figure}
User specific data is collected from social media platforms such as Facebook and Instagram with the users permission. These images are then undergo a series of pre processing tasks. This helps in cleaning the data. 

\begin{figure*}[thpb]
      \centering
        \includegraphics[width=0.75\textwidth]{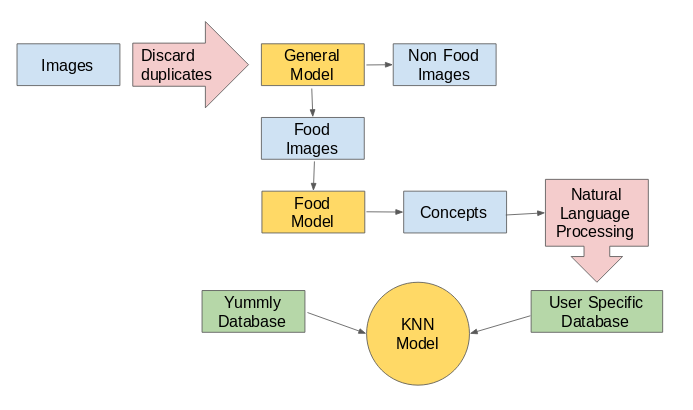}
      \caption{The above diagram represents the flow of the data pipeline along with the Models used.}
      
\end{figure*}

\section{METHODOLOGY}

The real task lies in converting the image into interpretable data that can be parsed and used. To help with this, a data processing pipeline is built.  The details of the pipeline are discussed below.
The data pipeline extensively uses the Clarifai\cite{GauravSood2017Clarifai:API} image recognition model. The 3 models used extensively are: 

\begin{enumerate}
\item The General Model : It recognizes over 11,000 different concepts and is a great all purpose solution. We have used this model to distinguish between Food images and Non-Food images.
\item The Food Model : It recognizes more than 1,000 food items in images down to the ingredient level. This model is used to identify the ingredients in a food image.
\item The General Embedding Model : It analyzes images and returns numerical vectors that represent the input images in a 1024-dimensional space. The vector representation is computed by using Clarifai’s ‘General’ model. The vectors of visually similar images will be close to each other in the 1024-dimensional space. This is used to eliminate multiple similar images of the same food item.
\end{enumerate}

\subsection{ DATA PRE PROCESSING}
\subsubsection{Distinctive Ingredients}

A cuisine can often be identified by some distinctive ingredients\cite{GhewariPredictingIngredients}. Therefore, we performed a frequency test to find the most occurring ingredients in each cuisine. Ingredients such as salt and water tend to show up at the top of these lists quite often but they are not distinctive ingredients. Hence, identification of unique ingredients is an issue that is overcome by individual inspection. For example:

\begin{table}[ht]
\caption{Unique Ingredients}
\label{table_example}
\begin{center}
\begin{tabular}{|c||c|}
\hline
Cuisine & Ingredients\\
\hline
Italian & Parmesan cheese, ricotta cheese\\
\hline
Mexican & corn tortillas, salsa\\
\hline
\end{tabular}
\end{center}
\end{table}
\subsubsection{To Classify Images as Food Images}

A dataset of 275 images of different food items from different cuisines was compiled. These images were used as input to the Clarifai Food Model. The returned tags were used to create a knowledge database. When the general model labels for an image with high probability were a part of this database, the image was classified as a food image.
The most commonly occurring food labels are visualized in Fig 3. 
\vspace{5 mm}
\begin{figure}[htbp]
      \centering
        \includegraphics[width=0.5\textwidth]{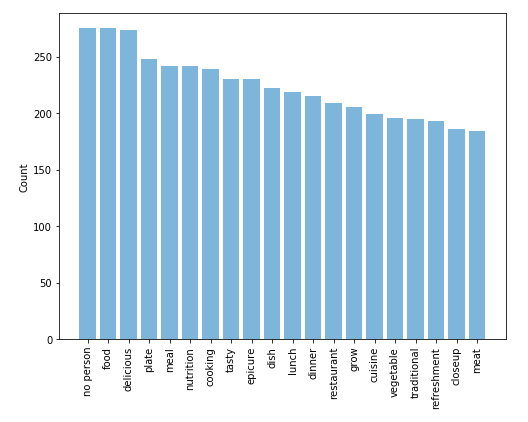}
        \caption{The top 20 most frequently occurring food labels}
\end{figure}

\subsubsection{To Remove Images with People}

To build a clean database for the user, images with people are excluded. This includes images with people holding or eating food. This is again done with the help of the descriptive labels returned by the Clarifai General Model. Labels such as "people" or "man/woman" indicate the presence of a person and such images are discarded.
\vspace{5 mm}
\subsubsection{To Remove Duplicate Images}

Duplicate images are removed by accessing the EXIF data of each image. Images with the same DateTime field are considered as duplicates and one copy is removed from the database. 
\vspace{5 mm}
\subsubsection{Natural Language Processing}

NLTK tools were used to remove low content adjectives from the labels/concepts returned from the Clarifai Models. This ensures that specific ingredient names are extracted without their unnecessary description. The Porter Stemmer Algorithm is used for removing the commoner morphological and inflectional endings from words.

\subsection{Basic Observations} 
From the food images(specific to each user), each image's descriptive labels are obtained from the Food Model. The Clarifai Food Model returns a list of concepts/labels/tags with corresponding probability scores on the likelihood that these concepts are contained within the image. The sum of the probabilities of each of these labels occurring in each image is plotted against the label in Fig 4.

\begin{figure}[htbp]
      \centering
        \includegraphics[width=0.5\textwidth]{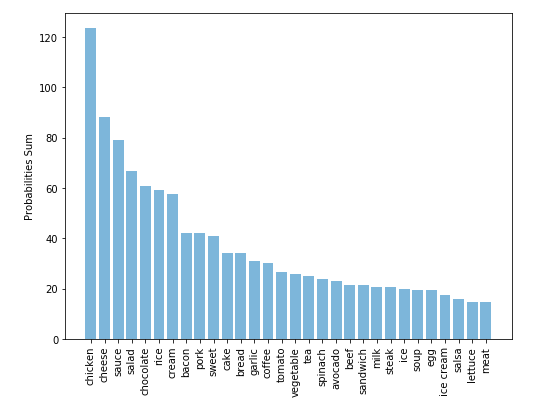}
        \caption{The sum of the probabilities of each label occurring in the images}
\end{figure}

The count of each of the labels occurring in each image is also plotted against each of the labels in Fig 5.
\begin{figure}[htbp]
      \centering
        \includegraphics[width=0.5\textwidth]{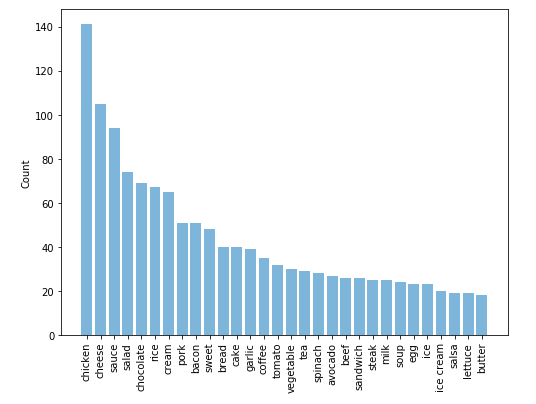}
        \caption{Count of each of the labels occurring in each image}
\end{figure}

\subsection{Rudimentary Method of Classification}

\begin{itemize}

\item Sometimes Clarifai returns the name of the dish itself. For example: "Tacos" which can be immediately classified as Mexican. There is no necessity to now map the ingredients to find the cuisine. Therefore, it is now necessary to maintain another database of native dishes from each cuisine. This database was built using the most popular or most frequently occurring dishes from each of the cuisines.
\item When no particular dish name was returned by the API, the ingredients with a probability of greater than 0.75 are selected from the output of the API. These ingredients are then mapped to the unique and frequently occurring ingredients from each cuisine. If more than 10 ingredients occur from a particular cuisine, the dish is classified into that cuisine. A radar map is plotted to understand the preference of the user. In this case, we considered only 10 cuisines.

\begin{figure}[htbp]
      \centering
        \includegraphics[width=0.5\textwidth]{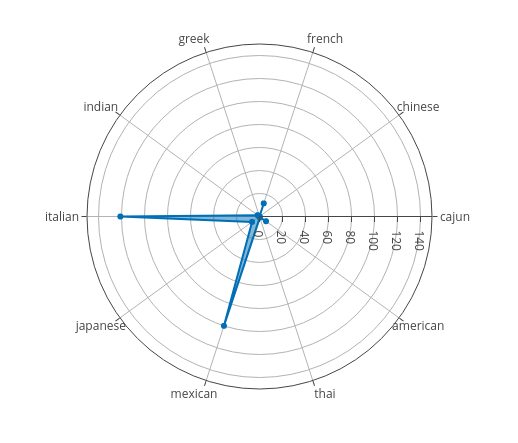}
        \caption{Radar chart depicting the frequency of predicted cuisines via the Rudimentary Method}
\end{figure}
\end{itemize}

\subsection{KNN Model for Classification}

A more sophisticated approach to classify based on the ingredients was adopted by using the K Nearest Neighbors Model. The Yummly dataset from Kaggle is used to train the model. The ingredients extracted from the images are used as a test set. The model was run successfully for k-values ranging from 1-25. The radar charts for some of the k values are shown in Fig 7, 8 and 9.

\begin{figure}[thpb]
      \centering
        \includegraphics[width=0.5\textwidth]{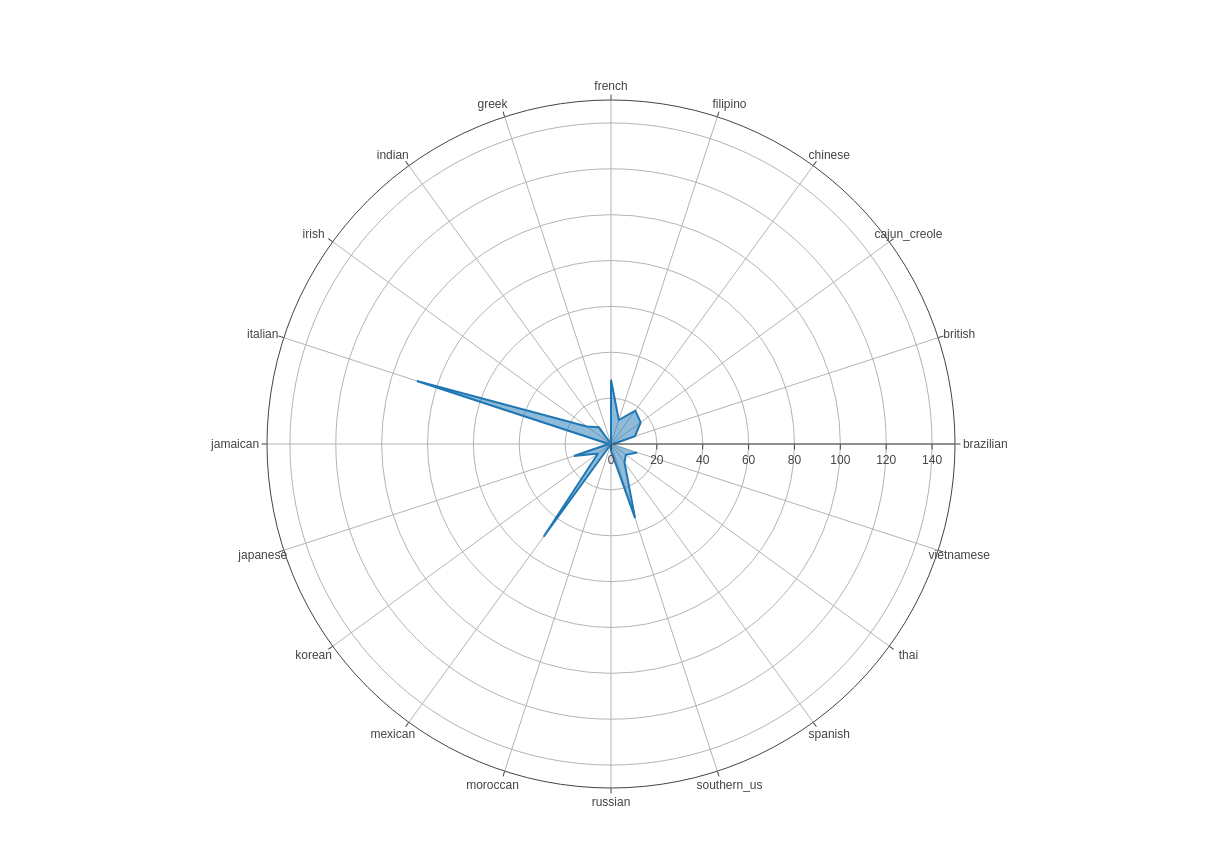}
      \caption{Radar chart indicating the frequency of cuisines in the prediction for K value = 2 i.e 2 nearest neighbors}
     
\end{figure}

\begin{figure}[thpb]
      \centering
        \includegraphics[width=0.5\textwidth]{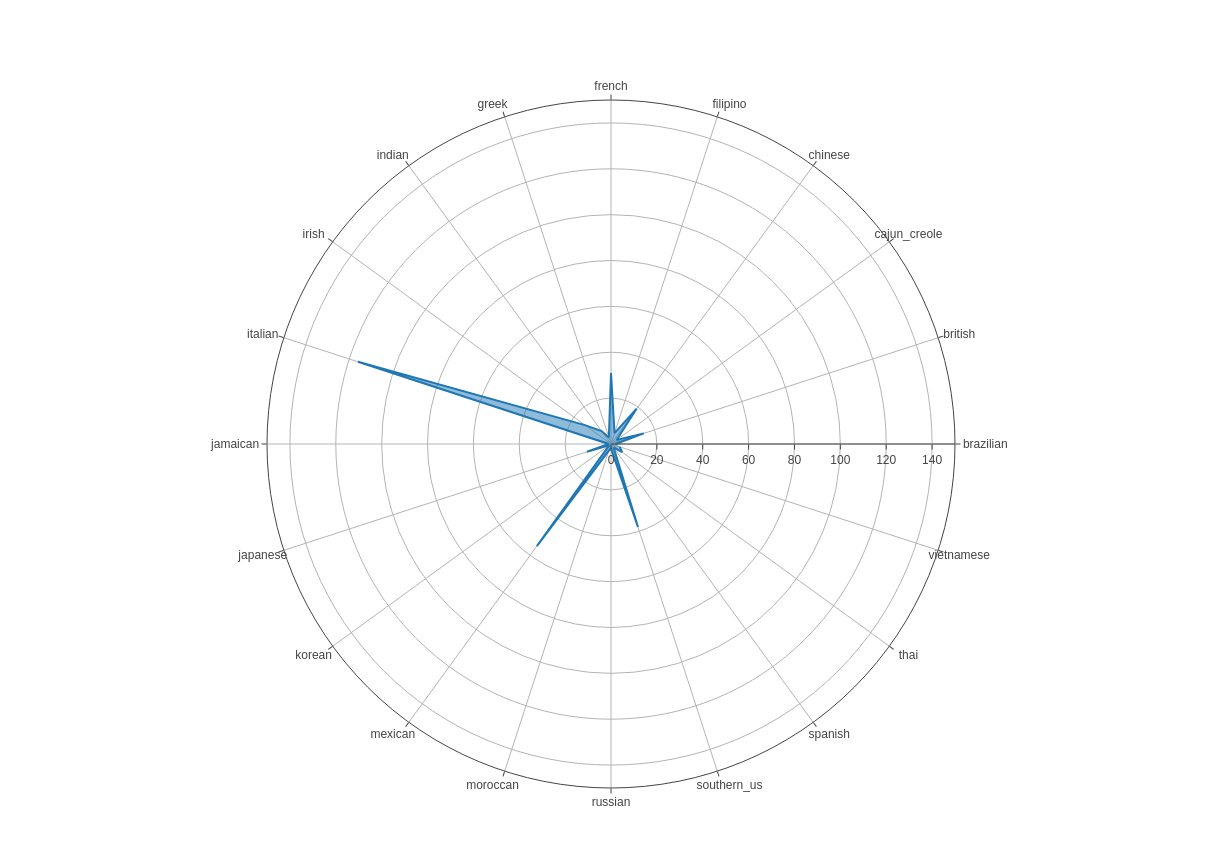}
      \caption{Radar chart indicating the frequency of cuisines in the prediction for K value = 10 i.e 10 nearest neighbors}
      
\end{figure}

\begin{figure}[thpb]
      \centering
        \includegraphics[width=0.5\textwidth]{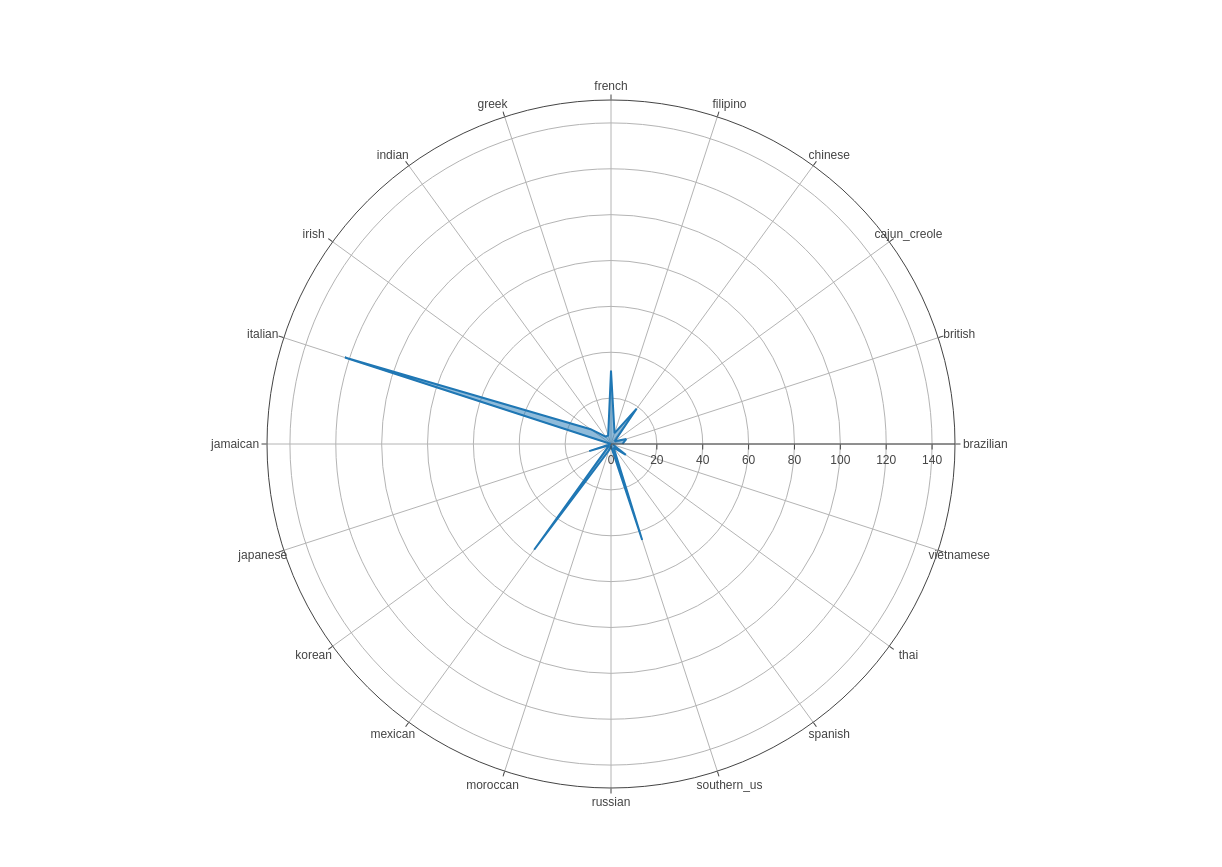}
      \caption{Radar chart indicating the frequency of cuisines in the prediction for K value = 20 i.e 20 nearest neighbors}
      
\end{figure}

Thus from these charts, we see that the user likes to eat Italian and Mexican food on most occasions. This is also in sync with the rudimentary method that we had used earlier.

\section{CONCLUSIONS}

In this paper, we present an effortless method to build a personal cuisine preference model. From images of food taken by each user, the data pipeline takes over, resulting in a visual representation of the user's preference. With more focus on preprocessing and natural text processing, it becomes important to realize the difficulty presented by the problem. We present a simple process to extract maximum useful information from the image.
We observe that there is significant overlap between the ingredients from different cuisines and the identified unique ingredients might not always be picked up from the image. Although, this similarity is what helps when classifying using the KNN model.
For the single user data used, we see that the 338 images are classified as food images. It is observed that Italian and Mexican are the most preferred cuisines. It is also seen that as K value increases, the number of food images classified into Italian increases significantly. Classification into cuisines like Filipino, Vietnamese and Cajun\_Creole decreases. This may be attributed to the imbalanced Yummly Dataset that is overshadowed by a high number of Italian recipes. 
\newline\textbf{Limitations :} The quality of the image and presentation of food can drastically affect the system. Items which look similar in shape and colour can throw the system off track. However, with a large database this should not matter much.
\newline\textbf{Future Directions :} The cuisine preferences determined for a user can be combined with the weather and physical activity of the user to build a more specific suggestive model. For example, if the meta data of the image were to be extracted and combined with the weather conditions for that date and time then we would be able to predict the type of food the user prefers during a particular weather. This would lead to a sophisticated recommendation system.

\addtolength{\textheight}{-12cm}   % This command serves to balance the column lengths
                                  % on the last page of the document manually. It shortens
                                  % the textheight of the last page by a suitable amount.
                                  % This command does not take effect until the next page
                                  % so it should come on the page before the last. Make
                                  % sure that you do not shorten the textheight too much.

%%%%%%%%%%%%%%%%%%%%%%%%%%%%%%%%%%%%%%%%%%%%%%%%%%%%%%%%%%%%%%%%%%%%%%%%%%%%%%%%

%%%%%%%%%%%%%%%%%%%%%%%%%%%%%%%%%%%%%%%%%%%%%%%%%%%%%%%%%%%%%%%%%%%%%%%%%%%%%%%%

%%%%%%%%%%%%%%%%%%%%%%%%%%%%%%%%%%%%%%%%%%%%%%%%%%%%%%%%%%%%%%%%%%%%%%%%%%%%%%%%

%%%%%%%%%%%%%%%%%%%%%%%%%%%%%%%%%%%%%%%%%%%%%%%%%%%%%%%%%%%%%%%%%%%%%%%%%%%%%%%%

\bibliographystyle{plain}
\bibliography{xyz}

\end{document}